\documentclass[a4paper,fleqn]{cas-dc}

\usepackage[authoryear]{natbib}
\usepackage{amssymb}
\usepackage{booktabs}
\usepackage{graphicx}
\usepackage{multirow}
\usepackage{placeins}
\usepackage{capt-of} 
\usepackage{needspace} 
\ExplSyntaxOn
\cs_new_eq:NN \ASCoROriginalCaptionof \captionof
\RenewDocumentCommand \captionof { m }
{
  \str_if_eq:nnT {#1} {figure}
  {
    \__reset_fig:
    \dim_set:Nn \l_fig_width_dim {\linewidth}
    \cs_set_eq:cN {@makecaption} \__make_fig_caption:nn
  }
  \ASCoROriginalCaptionof{#1}
}
\ExplSyntaxOff
\graphicspath{{./}{figures/}}

\def\tsc#1{\csdef{#1}{\textsc{\lowercase{#1}}\xspace}}
\tsc{WGM}
\tsc{QE}

\begin{document}
\let\WriteBookmarks\relax
\renewcommand{\topfraction}{.85}
\renewcommand{\dbltopfraction}{.85}
\renewcommand{\textfraction}{.08}
\renewcommand{\floatpagefraction}{.75}
\renewcommand{\dblfloatpagefraction}{.75}
\setlength{\dblfloatsep}{10pt plus 2pt minus 2pt}
\setlength{\dbltextfloatsep}{9pt plus 2pt minus 2pt}
\setlength{\textfloatsep}{9pt plus 2pt minus 2pt}
\setlength{\intextsep}{9pt plus 2pt minus 2pt}

\shorttitle{ASCoR: Semantic-Spatial Domain Generalization}
\shortauthors{Z. Wang et al.}

\title[mode=title]{ASCoR: Asymmetric Semantic-Spatial Learning for Remote Sensing Domain Generalization}

\author[1]{Zitong Wang}[type=author]



\ead{23371278@buaa.edu.cn}



\affiliation[1]{organization={Beihang University},
            addressline={Xueyuan Road No.37, Haidian district}, 
            city={Beijing},
            postcode={100191}, 
            state={},
            country={China}}

\affiliation[2]{organization={University of Science and Technology Beijing},
            addressline={Xueyuan Road No.30, Haidian district}, 
            city={Beijing},
            postcode={100083}, 
            state={},
            country={China}}

\affiliation[3]{organization={University of Liverpool},
            addressline={Foundation Building,
Brownlow Hill,}, 
            city={Liverpool},
            postcode={L693BX}, 
            state={},
            country={UK}}

\affiliation[4]{organization={Guangdong Electronic Information Engineering Research Institute, University of Electronic Science and Technology of China},
            addressline={Zongbu 2nd Road No. 17}, 
            city={Dongguan},
            postcode={523808}, 
            state={},
            country={China}}

\author[2]{Shuchang Lyu}[type=author, orcid=0000-0001-9769-7083]
\cormark[1]


\ead{lyushuchang@ustb.edu.cn}

\author[1]{Meng Li}[type=author]


\ead{limenglm@buaa.edu.com}


\author[4]{Qiang Wang}[type=author]
\ead{wq@cap.edu.cn}

\author[3]{Guangliang Cheng}[type=author]
\ead{Guangliang.Cheng@liverpool.ac.uk}

\author[1]{Zhenwei Shi}[type=author]
\ead{shizhenwei@buaa.edu.cn}

\cortext[1]{Corresponding author}

\begin{abstract}
Source-only domain generalization for remote sensing semantic segmentation aims to achieve reliable prediction in unseen domains without accessing target-domain data during training. Vision foundation models offer transferable knowledge for this task, but its effective use faces two challenges. Category recognition remains vulnerable to source-specific visual cues. Learning task-specific dense representations also risks excessive source-domain specialization, potentially weakening pretrained transferability. To solve these problems, we propose ASCoR, an asymmetric semantic--spatial learning framework built on SAM3. We exploit SAM3's concept-aware segmentation priors to provide explicit category guidance beyond source-domain visual observations. Specifically, the text-conditioned semantic branch associates dense predictions with class-name prompts. Its Probe-Once Dynamic Adaptive LoRA allocates a constrained adaptation budget across encoder blocks to control source-specific specialization. A complementary task-specific dense branch learns multi-scale spatial representations upon a separate frozen SAM3 image encoder. The two branches are integrated through a lightweight reliability router that applies bounded spatial correction according to local prediction reliability. The entire framework requires only source-domain supervision. Extensive experiments on the ISPRS, LoveDA, and OpenEarthMap benchmark datasets demonstrate the effectiveness of ASCoR across diverse domain shifts. On six ISPRS transfers, ASCoR achieves a state-of-the-art average mIoU of 68.61\%, surpassing the previous best method by 2.69 percentage points. It further achieves the best performance on all three LoveDA/OpenEarthMap transfers, reaching 59.07\%, 47.95\%, and 49.38\% mIoU on Rural-to-Urban, Urban-to-Rural, and Urban-to-OpenEarthMap, respectively. Systematic ablation studies assess the contribution of each component. Feature-space visualizations and spatial reliability analyses provide further insight into the framework's cross-domain representations and complementary correction mechanism. The code is available at \url{https://github.com/WangZitong42/ASCoR}.
\end{abstract}





\maketitle
\flushbottom
\setlength{\abovedisplayskip}{8pt plus 2pt minus 2pt}
\setlength{\belowdisplayskip}{8pt plus 2pt minus 2pt}
\setlength{\abovedisplayshortskip}{4pt plus 2pt minus 1pt}
\setlength{\belowdisplayshortskip}{6pt plus 2pt minus 2pt}
\section{Introduction}
\label{sec:introduction}

Remote sensing semantic segmentation provides pixel-level land-cover information for applications such as urban planning and environmental monitoring~\citep{zhu2017deep,ma2019deep}. However, models trained on one domain often lose accuracy when deployed elsewhere because differences in geographic regions, imaging sensors, and scene compositions can substantially alter the visual distribution~\citep{wang2021loveda,iizuka2023frequency,liang2024single}. This gap between source-domain learning and practical deployment motivates source-only domain generalization (DG), which aims to transfer a model trained on a single labeled source domain to unseen domains without accessing target-domain data during training~\citep{zhou2022domain,rafi2024domain}.

To facilitate such cross-domain transfer, vision foundation models (VFMs), which acquire transferable knowledge through large-scale pretraining, offer a promising basis for extending remote sensing representations beyond the visual regularities of a single source domain~\citep{he2022masked,oquab2023dinov2,mendieta2023towards}. Their relevance to dense prediction becomes more evident as pretrained knowledge encompasses explicit semantic concepts and structural priors: vision-language models associate visual observations with category semantics~\citep{radford2021learning}, whereas segmentation foundation models capture object-level spatial structure~\citep{kirillov2023segment}. By incorporating concept recognition into promptable segmentation, SAM3 connects these two capabilities, creating an opportunity to guide dense land-cover prediction with explicit category semantics~\citep{carion2025sam}.

\begin{figure*}[!tbp]
  \centering
  \includegraphics[width=\textwidth]{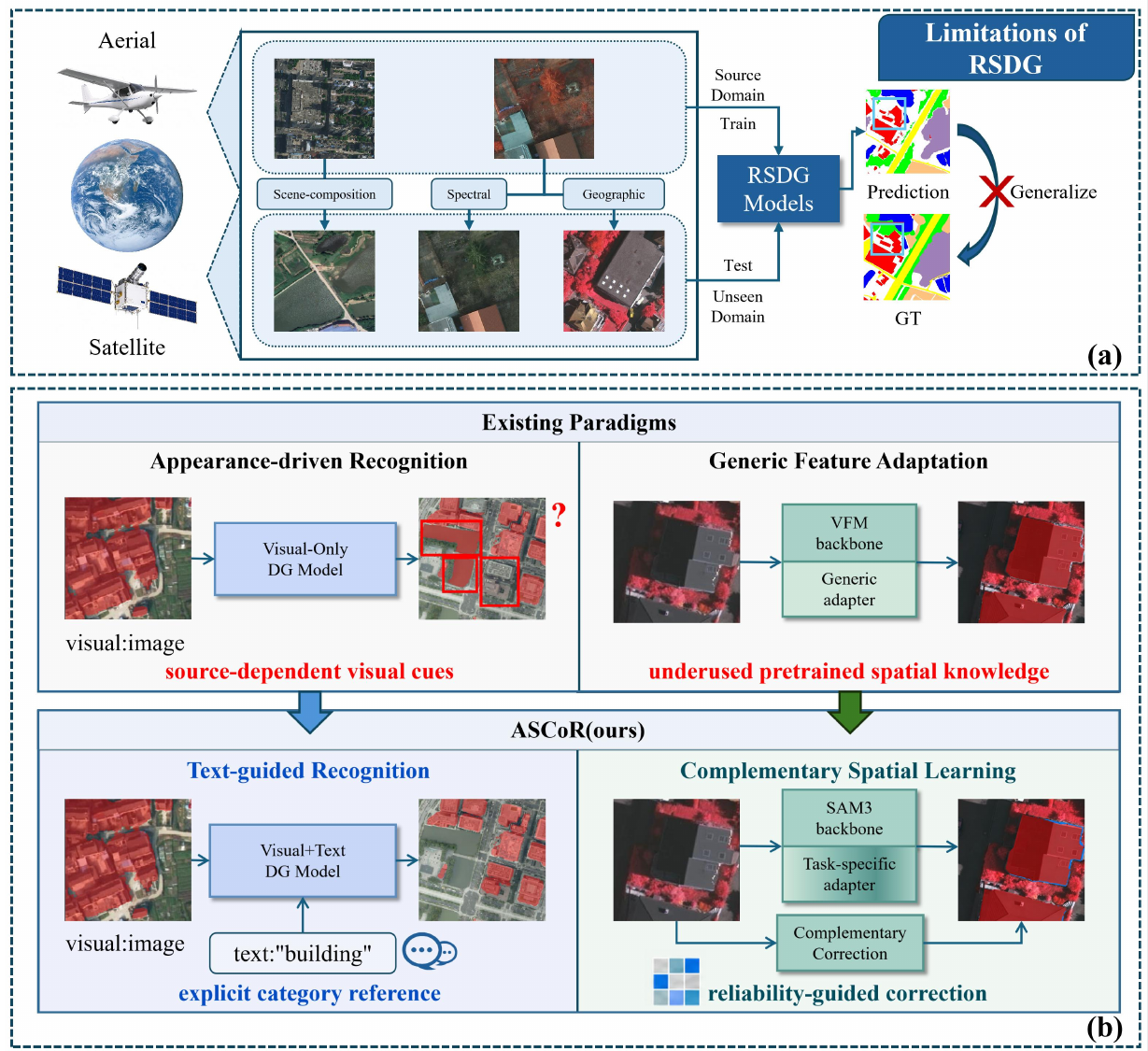}
  \caption{Motivation and paradigm comparison for source-only remote sensing domain generalization. (a) Limitations of representative visual-driven DG approaches in transferring category semantics and exploiting task-specific dense representations. (b) ASCoR combines text-conditioned semantic guidance with complementary dense spatial learning through an asymmetric SAM3 framework and reliability-guided correction. Earth image credit: NASA/Apollo 17, via Wikimedia Commons (public domain).}
  \label{fig:scientific_problem}
\end{figure*}

Much of the existing research on remote sensing DG focuses on conventional segmentation networks, whose robustness is improved through appearance diversification or representation regularization~\citep{iizuka2023frequency,liang2024single,wang2025boosting}. Although these strategies address visual variability, category discrimination remains primarily dependent on source-domain visual observations, leaving predictions vulnerable to changes in class appearance. A smaller body of work introduces VFMs to obtain broader pretrained knowledge through the adaptation of generic visual representations~\citep{gong2025crossearth,zhang2026spectralx}. Within this paradigm, the potential for extracting task-specific spatial evidence that complements category semantics remains insufficiently explored. As illustrated in Fig.~\ref{fig:scientific_problem}(a), these limitations highlight the need for explicit category guidance and a more effective use of pretrained representations for dense spatial prediction.

To solve these problems, we propose ASCoR, an asymmetric semantic--spatial learning framework built on SAM3. Figure~\ref{fig:scientific_problem}(b) contrasts our framework with the representative visual-driven DG paradigm. To strengthen category discrimination beyond source-domain visual cues, we exploit SAM3's concept-aware segmentation capability to construct a text-conditioned semantic branch (TCSB). By associating dense predictions with class-name prompts, TCSB introduces an explicit semantic reference for recognizing land-cover categories across variations in visual appearance.

To further exploit task-specific spatial information, we develop a complementary task-specific dense branch (TSDB) upon a separate SAM3 encoder whose pretrained weights remain frozen. Multi-scale dense decoding extracts spatially detailed representations from hierarchical features, supported by bounded structure injection that incorporates boundary and geometric cues. Complementary supervision from TCSB directs this learning toward regions where its semantic predictions are less reliable. A lightweight reliability router then estimates the local utility of the dense predictions and applies bounded spatial correction. Together, these components enable task-specific spatial evidence to supplement category-level semantic guidance.

We further introduce Probe-Once Dynamic Adaptive LoRA to control adaptation capacity within TCSB. By allocating a constrained low-rank budget across encoder blocks according to their source-task adaptation responses, this strategy supports task-specific learning with limited modification of pretrained representations. The complete framework is optimized exclusively with source-domain supervision.

Extensive experiments on the ISPRS, LoveDA, and OpenEarthMap benchmark datasets demonstrate the effectiveness of ASCoR across nine transfers. ASCoR achieves a state-of-the-art average mIoU of 68.61\% on six ISPRS transfers, exceeding the previous best method by 2.69 percentage points. It also achieves the best performance on all three LoveDA/OpenEarthMap transfers. Systematic ablations establish the contributions of the proposed components, supported by feature visualizations and spatial reliability analyses that provide insight into their complementary behavior.

The main contributions of this work are summarized as follows:

\begin{itemize}

    \item To the best of our knowledge, we are the first to introduce SAM3-based text-conditioned semantic modeling into source-only domain generalization for remote sensing semantic segmentation, providing explicit category guidance beyond source-domain visual observations.

    \item We develop an asymmetric semantic--spatial learning framework with reliability-guided fusion to better exploit task-specific dense representations and strengthen their complementarity with category-level semantic guidance.

    \item We propose Probe-Once Dynamic Adaptive LoRA to control source-specific adaptation under a constrained parameter budget, supporting effective task learning with limited modification of pretrained representations.

\end{itemize}

\section{Related Work}
\label{sec:related_work}

\subsection{Remote Sensing Semantic Segmentation and Domain Generalization}

Remote sensing semantic segmentation assigns pixel-level land-cover labels to aerial or satellite imagery and serves as a fundamental technique for land-cover mapping, urban planning, agricultural monitoring, and environmental analysis~\citep{zhu2017deep,ma2019deep}. CNN-based methods established dense prediction, encoder-decoder learning, and multi-scale contextual modeling as standard segmentation paradigms~\citep{long2015fully,ronneberger2015u,chen2018encoder,zhao2017pyramid}. More recently, Transformer- and mask-classification-based architectures further improved long-range representation learning and mask-level prediction~\citep{xie2021segformer,xiao2018unified,cheng2021per,cheng2022masked}.

Despite these advances, remote sensing segmentation remains highly sensitive to domain shifts caused by geographic regions, sensors, spatial resolutions, seasons, and acquisition conditions~\citep{wang2021loveda,iizuka2023frequency,liang2024single}. Such shifts may affect not only image appearance but also object scale, spatial layout, land-cover composition, and structural patterns. Domain adaptation alleviates this problem by exploiting target-domain data during training through adversarial alignment, output-space adaptation, self-training, or pseudo-label refinement~\citep{ganin2015unsupervised,tsai2018learning,hoffman2018cycada,zou2019confidence,vu2019advent}. Remote sensing domain adaptation has likewise been investigated through curriculum learning, adaptive discrimination, and cross-city semantic alignment~\citep{zhang2021curriculum,chen2022unsupervised,hong2023cross}.

Domain generalization (DG), in contrast, trains models only on source-domain data and directly evaluates them on unseen target domains~\citep{muandet2013domain,li2018learning,balaji2018metareg,kim2021selfreg,zhou2022domain,rafi2024domain}. For remote sensing segmentation, FOSMix enhances robustness through frequency-based style mixing~\citep{iizuka2023frequency}, CCDR constructs auxiliary domains using texture and style randomization while enforcing category consistency~\citep{liang2024single}, and SMGPC combines style mapping with prototype-based contrastive learning for category-level generalization~\citep{wang2025boosting}. These methods demonstrate the effectiveness of visual-domain diversification and representation regularization. Nevertheless, explicit category semantics beyond source-domain visual observations remain relatively underexplored, particularly for source-only dense prediction under complex remote sensing shifts.

\subsection{Vision-Language and Foundation Models for Remote Sensing}

Large-scale pretraining has substantially improved the transferability of visual representations. Self-supervised models such as MAE, DINO, and DINOv2 learn generic visual features from large image collections~\citep{he2022masked,caron2021emerging,oquab2023dinov2}, while CLIP aligns visual representations with natural-language concepts through large-scale image-text supervision~\citep{radford2021learning}. This vision-language paradigm has also been extended to dense prediction. LSeg aligns pixel-level visual embeddings with text embeddings for language-driven semantic segmentation~\citep{li2022language}, while DenseCLIP transfers CLIP knowledge from image-text matching to pixel-text matching for dense prediction~\citep{rao2022denseclip}. These studies demonstrate the potential of linguistic semantics to provide explicit category-level guidance beyond purely visual representation learning.

Foundation-model pretraining has also been actively explored for remote sensing image understanding. Continual pretraining has been used to construct geospatial foundation representations~\citep{mendieta2023towards}, while Scale-MAE and SatMAE incorporate scale, temporal, and multispectral characteristics into remote sensing pretraining~\citep{reed2023scale,cong2022satmae}. SpectralGPT and HyperSIGMA further extend foundation representation learning to spectral remote sensing imagery~\citep{hong2024spectralgpt,wang2025hypersigma}. More recently, CrossEarth demonstrates the potential of geospatial foundation models for domain-generalized remote sensing semantic segmentation~\citep{gong2025crossearth}.

Although these studies provide increasingly transferable representations, generic foundation features do not necessarily offer the category-aware semantics and dense spatial representations required by a specific downstream segmentation task. In particular, how to exploit linguistic semantic knowledge together with generic foundation representations for source-only remote sensing domain generalization remains insufficiently explored. Our work addresses this gap by introducing explicit category text into domain-generalized remote sensing semantic segmentation and jointly exploiting category-aware semantic guidance and task-specific dense representations.

\subsection{SAM-style Foundation Models for Remote Sensing Interpretation}

The Segment Anything Model (SAM) learns promptable, class-agnostic segmentation priors from large-scale segmentation data~\citep{kirillov2023segment}. Its object-level and structural representations provide a foundation for delineating land-cover objects, motivating efforts to transfer these capabilities to remote sensing imagery.

Such efforts have explored both data construction and model adaptation. SAMRS employs SAM-generated masks to scale up remote sensing segmentation data~\citep{wang2023samrs}, while RingMo-SAM extends SAM to multimodal remote sensing imagery~\citep{yan2023ringmo}. RSPrompter introduces learnable prompting for remote sensing instance segmentation~\citep{chen2024rsprompter}, and MeSAM enhances SAM with multi-scale adaptation for optical remote sensing imagery~\citep{zhou2024mesam}. These studies establish the utility of general segmentation priors for remote sensing interpretation.

More recently, SAM3 has extended promptable segmentation to concept-aware recognition~\citep{carion2025sam}, opening further opportunities for category-level remote sensing interpretation. SegEarth-OV3 explores this capability through training-free open-vocabulary segmentation, combining mask outputs and presence-based category filtering~\citep{li2025segearth}. ProC-SAM3 further develops prompt calibration and presence-guided fusion for remote sensing open-vocabulary semantic segmentation~\citep{11609477}. These studies demonstrate the potential of concept-aware priors for training-free prediction. Applying such priors to source-only domain generalization introduces a complementary challenge: learning task-specific semantic and spatial representations from a single labeled source domain while preserving transferability to unseen domains with different geographic and imaging conditions.

\subsection{Parameter-Efficient Adaptation and Reliability-Guided Fusion}

Parameter-efficient fine-tuning (PEFT) adapts pretrained models by updating only a small set of task-specific parameters while keeping most foundation weights frozen. Representative approaches include Adapter tuning, Visual Prompt Tuning, LoRA, and AdaptFormer~\citep{houlsby2019parameter,jia2022visual,hu2022lora,chen2022adaptformer}. Such strategies are particularly attractive for large foundation models because they reduce adaptation cost while limiting direct modification of pretrained parameters.

Recent work further investigates adaptive low-rank allocation. AdaLoRA dynamically reallocates a limited low-rank budget according to parameter importance~\citep{zhang2023adalora}, whereas DyLoRA trains LoRA modules to operate across a range of ranks without repeated rank-specific retraining~\citep{valipour2023dylora}. In remote sensing, SpectralX demonstrates the effectiveness of parameter-efficient adaptation for domain-generalized spectral foundation models~\citep{zhang2026spectralx}. Meanwhile, full fine-tuning has been shown to distort pretrained features and degrade out-of-distribution performance under large distribution shifts~\citep{kumar2022fine}. These observations motivate controlled foundation-model adaptation for source-only DG.

Different from existing adaptive LoRA methods developed primarily for general pretrained models, our Dynamic Adaptive LoRA targets hierarchical SAM3 adaptation under a constrained total rank budget. It allocates adaptation capacity according to the contribution of different encoder layers to downstream remote sensing semantic learning, aiming to preserve transferable foundation knowledge while acquiring task-specific representations.

Finally, the adapted semantic branch and frozen dense branch provide complementary information but may exhibit different reliability across categories and unseen domain shifts. A fixed fusion rule implicitly assumes stable branch quality and may therefore be suboptimal. We instead employ reliability-guided class-wise fusion to dynamically integrate category-aware semantic priors and task-specific dense predictions according to their prediction reliability.

\section{Methodology}
\label{sec:method}

\subsection{Overview}
\label{subsec:overview}

ASCoR is trained on a single labeled source domain and evaluated directly on unseen target domains. Each input is a three-channel remote sensing image with pixel-wise land-cover annotations. The framework combines explicit category guidance with task-specific spatial evidence using only source-domain supervision. Target-domain images are not used for parameter optimization.

The framework consists of a text-conditioned semantic branch (TCSB), a task-specific dense branch (TSDB), and a reliability-guided fusion module. The branches have distinct learning objectives: TCSB uses category prompts to guide semantic discrimination, whereas TSDB extracts spatial evidence targeted at the limitations of the semantic prediction. Their pretrained SAM3 weights remain frozen, with task learning concentrated in the respective adaptation and decoding components. The router subsequently determines where and to what extent the dense evidence should correct TCSB. Superscripts $A$ and $B$ denote TCSB and TSDB, respectively; $\mathbf{Z}$ denotes logits, and $\mathbf{P}$ denotes the corresponding class probabilities obtained through softmax.

\begin{figure*}[!tbp]
  \centering
  \includegraphics[width=\textwidth]{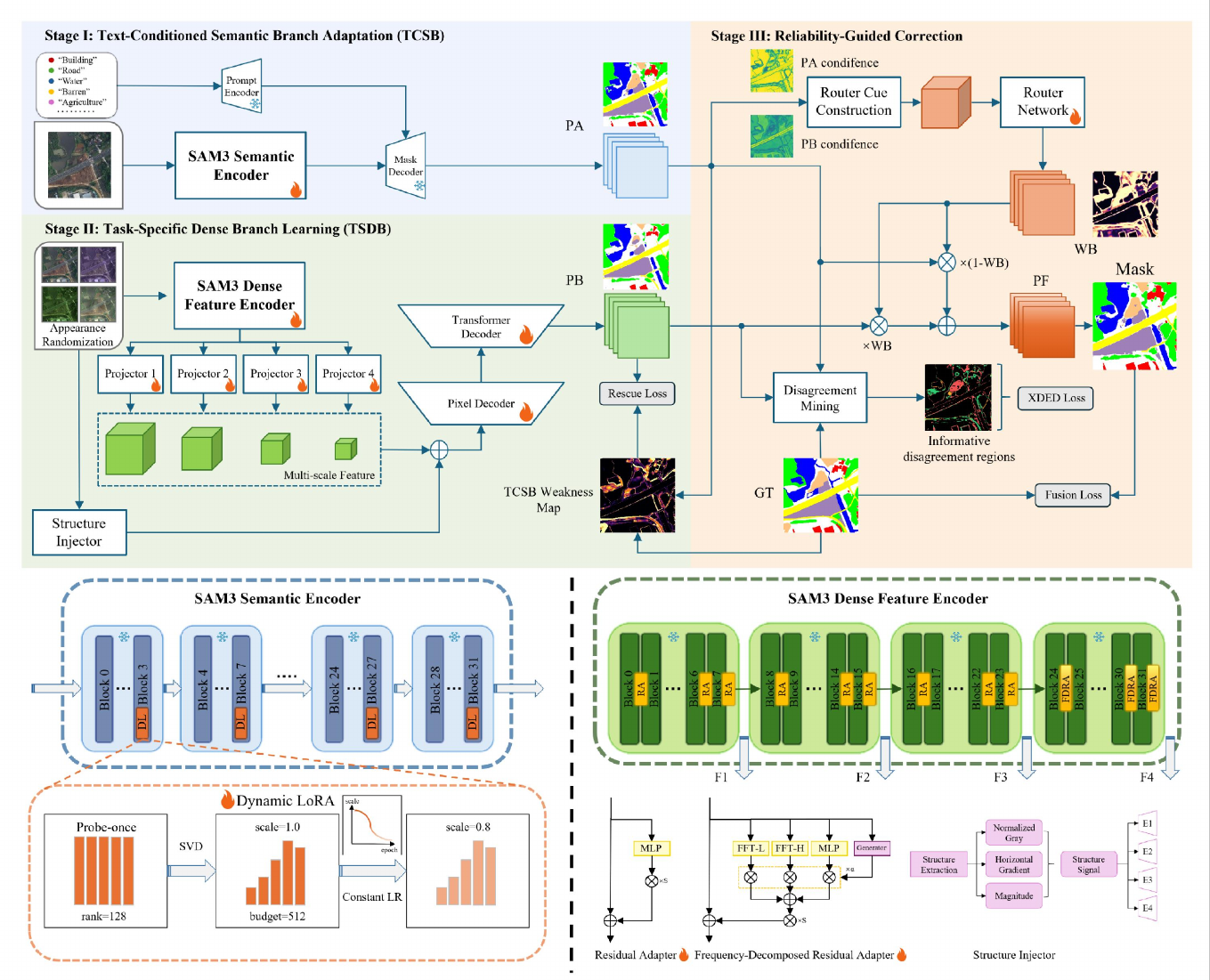}
  \caption{Architecture and three-stage training of ASCoR. TCSB combines category prompts with Probe-Once Dynamic Adaptive LoRA. TSDB learns complementary spatial representations through trainable adaptation, structure, and decoding modules upon frozen pretrained SAM3 weights. After both branches are frozen, the reliability router learns bounded spatial correction using source-domain supervision.}
  \label{fig:framework}
\end{figure*}

Figure~\ref{fig:framework} summarizes the architecture and three-stage training procedure.

\subsection{Text-Conditioned Semantic Branch}
\label{subsec:tcsb}

TCSB introduces class-name prompts into SAM3 so that category prediction is guided by explicit semantic concepts as well as visual observations. A fixed set of land-cover concepts is used throughout each transfer task, providing a common semantic reference across changes in image appearance. Each prompt produces a dense class response, and these responses are combined into the segmentation logits. For an unprompted background or clutter category, the response is set to the negative maximum foreground response at each pixel.

To adapt SAM3 with controlled capacity, the pretrained weights remain frozen and trainable LoRA residuals are introduced into selected encoder blocks~\citep{hu2022lora}. Omitting transformation indices, the forward computation is
\begin{equation}
    \mathbf{h}'=\mathbf{W}\mathbf{h}
    +g(e)\lambda\,\mathbf{B}_{r_l}\mathbf{A}_{r_l}
    \mathcal{D}(\mathbf{h}),
\label{eq:lora_residual}
\end{equation}
where $\mathbf{W}$ is the frozen weight, $\mathcal{D}$ denotes dropout, and $\mathbf{A}_{r_l}$ and $\mathbf{B}_{r_l}$ are the active factor slices assigned to block $l$. The coefficient $\lambda$ scales the local residual, and $g(e)$ controls its global strength.

Uniform ranks do not account for differences in adaptation response across encoder blocks. We therefore use a short source-domain probe to determine how the available capacity should be distributed. All candidate blocks are first adapted with a common rank $r_p$, and the energy of their learned updates provides a source-task response score. For block $l$ containing $M_l$ adapted transformations, the score is
\begin{equation}
    s_l=\frac{1}{M_l}\sum_{m=1}^{M_l}
    \left\|\mathbf{B}^{(p)}_{l,m}
    \mathbf{A}^{(p)}_{l,m}\right\|_F^2.
\label{eq:rank_score}
\end{equation}
A fixed total budget $B$ is distributed across the $L$ candidate blocks:
\begin{equation}
\begin{aligned}
    \widetilde r_l
    &=r_{\min}+(B-Lr_{\min})
      \frac{s_l}{\sum_{j\in\mathcal{L}}s_j},\\
    \mathbf{r}^{*}
    &=\Pi_{\mathcal{C}}(\widetilde{\mathbf{r}}).
\end{aligned}
\label{eq:rank_allocation}
\end{equation}
Here, $\mathcal{L}$ is the candidate-block set. The projection $\Pi_{\mathcal{C}}$ produces integer ranks whose sum equals $B$, with each rank bounded by $r_{\min}$ and $r_{\max}$. Stronger probe responses therefore receive more adaptation capacity without increasing the total rank budget. Adaptive-rank PEFT methods have explored importance-based allocation and multi-rank learning~\citep{zhang2023adalora,valipour2023dylora}. ASCoR performs allocation once after probing, separating capacity estimation from final task learning and keeping the active rank structure fixed thereafter.

Changing the rank map also requires transferring the updates learned during probing. Simply truncating the two LoRA factors independently can discard useful directions because the effective update is determined by their product. We therefore apply compact SVD to the probed update and retain the leading singular components permitted by both the probe rank and the newly allocated rank. Let $\mathbf{U}_k$, $\mathbf{\Sigma}_k$, and $\mathbf{V}_k$ denote these retained components. The retained factor initialization is
\begin{equation}
\begin{aligned}
    \mathbf{B}^{*}_k
    &=\mathbf{U}_k\mathbf{\Sigma}_k^{1/2},\\
    \mathbf{A}^{*}_k
    &=\mathbf{\Sigma}_k^{1/2}\mathbf{V}_k^{\top}.
\end{aligned}
\label{eq:svd_factorization}
\end{equation}
These directions are transferred into the allocated factor tensors before final adaptation. The global residual strength then follows cosine decay from its initial value of one to a prescribed minimum. TCSB is optimized with standard pixel-wise cross-entropy over valid source labels.

\subsection{Task-Specific Dense Branch}
\label{subsec:tsdb}

TSDB is designed to learn spatial information that supplements the text-conditioned prediction. Its pretrained SAM3 encoder weights remain frozen, while lightweight adaptation, feature projection, structure injection, and dense decoding modules are trained. Frequency-aware adaptation provides adjustment within the feature hierarchy, and selected hierarchical features are projected into a common embedding space and spatially aligned so that the decoder can combine information from multiple representation levels.

Category-level guidance alone does not explicitly determine the spatial detail required by dense land-cover prediction. Before decoding, TSDB therefore supplements the projected features with geometric cues derived from the input image. A lightweight structure encoder processes normalized grayscale and Sobel-derived edge information, providing boundary evidence alongside the learned appearance features. Denoting this structural input by $\mathbf{S}$ and the projected feature at level $k$ by $\mathbf{P}_k$, the injection is
\begin{equation}
    \widehat{\mathbf{P}}_k
    =\mathbf{P}_k+\tanh(a_k)\mathcal{J}_k(\mathbf{S}),
\label{eq:structure_injection}
\end{equation}
where $\mathcal{J}_k$ combines scale alignment with structure encoding. The learnable coefficient is bounded by the hyperbolic tangent, allowing the contribution of structural cues to be regulated separately at each feature level. The resulting features are processed by a query-based decoder following Mask2Former~\citep{cheng2022masked}. Its query predictions jointly encode class information and spatial masks; class-weighted masks are aggregated into dense class scores and converted to logits by a logarithm.

TSDB is trained using original source images and geometry-preserving randomized views. The appearance perturbations simulate spectral, atmospheric, sensor, and frequency variations without changing pixel correspondence, allowing the original annotations to supervise both views. Training on these views exposes the dense branch to alternative appearances of the same labeled structures while retaining direct supervision on the original source images.

To encourage complementarity, the frozen TCSB additionally directs TSDB learning toward pixels that it handles less reliably. The weighting uses TCSB's probability for the ground-truth category, so pixels receiving an incorrect high-confidence prediction can still obtain strong complementary supervision. For a randomized source image, the weight at pixel $p$ is
\begin{equation}
    \omega_p=\operatorname{clip}\big[
    (1-P^A_{y_p,p})^\gamma,\,0,\,\omega_{\max}\big],
\label{eq:rescue_weight}
\end{equation}
where $P^A_{y_p,p}$ is TCSB's probability for the ground-truth category on that randomized view. Pixels assigned a high probability for the correct category receive less additional emphasis. The complementary objective is a normalized weighted loss:
\begin{equation}
    \mathcal{L}_{\mathrm{res}}
    =\frac{\sum_{p\in\Omega}\omega_p
    \ell_{\mathrm{CE}}(\mathbf{Z}^B_p,y_p)}
    {\sum_{p\in\Omega}\omega_p+\epsilon},
\label{eq:rescue_loss}
\end{equation}
where $\Omega$ contains valid source pixels and $\ell_{\mathrm{CE}}$ is pixel-wise cross-entropy. Predictions in this objective are evaluated on the randomized view. The weight cap controls the maximum pixel-wise emphasis, and normalization makes the objective depend on the relative distribution of complementary weights.

Let $\mathcal{L}_M^o$ and $\mathcal{L}_M^a$ denote the standard query-decoder segmentation losses on the original and randomized views, respectively. The complete TSDB objective is
\begin{equation}
    \mathcal{L}_B
    =\frac{\mathcal{L}_M^o+
    \lambda_{\mathrm{aug}}\mathcal{L}_M^a}
    {1+\lambda_{\mathrm{aug}}}
    +\lambda_{\mathrm{res}}(e)\mathcal{L}_{\mathrm{res}}.
\label{eq:tsdb_loss}
\end{equation}
The augmentation coefficient balances supervision across the two views, whose segmentation losses form a weighted average. The complementary coefficient gradually increases during early training, introducing greater emphasis on TCSB's weak regions as dense prediction develops. TSDB is therefore trained both to perform the segmentation task and to acquire evidence that can supplement the semantic branch.

\subsection{Reliability-Guided Complementary Fusion}
\label{subsec:fusion}

The two branches provide different evidence whose relative reliability varies across categories and spatial locations. A branch with lower overall accuracy may still supply useful corrections in particular regions, so the contribution of TSDB should depend on local prediction quality. The router uses prediction-based reliability cues to estimate this complementary contribution.

Its input $\mathbf{R}$ concatenates both probability maps, their absolute difference, one-hot category predictions, branch-wise confidence, normalized entropy, top-two probability margins, and the binary disagreement map. The class probabilities and category predictions retain semantic information; the uncertainty statistics describe the strength of each prediction, and the difference and disagreement cues identify potential corrections. Together, they allow routing to depend on both category identity and local reliability without requiring target labels.

A lightweight spatial network produces a sigmoid gate and applies bounded correction to the primary semantic prediction:
\begin{equation}
\begin{aligned}
    g_p
    &=\sigma\!\left(\mathcal{G}_{\psi}(\mathbf{R})_p\right),\\
    \mathbf{P}^F_p
    &=\mathbf{P}^A_p+
      \beta g_p(\mathbf{P}^B_p-\mathbf{P}^A_p).
\end{aligned}
\label{eq:bounded_fusion}
\end{equation}
Here, $\beta$ sets the maximum contribution of the dense branch. TCSB remains the primary estimate, and TSDB supplies a spatially varying residual correction. The gate allows the amount of dense evidence to vary across the image instead of imposing the same mixing weight everywhere. Its bounded contribution maintains the intended asymmetric roles of the two branches.

Router supervision measures the value of a correction using source labels. Disagreement alone does not establish that TSDB should be trusted, since either branch may be incorrect. We therefore supervise the gate according to whether TSDB increases the probability assigned to the ground-truth category. At valid disagreement pixels, the relative utility target is
\begin{equation}
    u_p=
    \frac{[P^B_{y_p,p}-P^A_{y_p,p}]_+}
    {\max(P^B_{y_p,p},\epsilon)}.
\label{eq:soft_utility}
\end{equation}
The positive-part operator sets the target to zero when TSDB offers no correct-class probability improvement. Otherwise, the target increases continuously with its relative advantage. This provides graded supervision for the correction gate based on the local relationship between the two predictions.

To encourage stable routing under appearance changes, utility learning is combined with gate consistency across geometry-preserving source views. Since these views retain pixel correspondence, their gate responses can be compared at aligned locations:
\begin{equation}
\begin{aligned}
    \mathcal{L}_{\mathrm{util}}
    &=\mathbb{E}_{p\sim\Omega_d}
      [\operatorname{BCE}(g_p,u_p)],\\
    \mathcal{L}_{\mathrm{cons}}
    &=\mathbb{E}_{p\sim\Omega_u}
      [\operatorname{SmoothL1}(g_p,\widetilde g_p)].
\end{aligned}
\label{eq:router_auxiliary}
\end{equation}
Here, $\Omega_d$ contains valid disagreement pixels, $\Omega_u$ is the valid cross-view consistency region, and $\widetilde g_p$ is the gate from the randomized counterpart. The fused prediction also receives standard pixel-wise cross-entropy supervision, denoted by $\mathcal{L}_{\mathrm{fuse}}$. The complete router objective is
\begin{equation}
    \mathcal{L}_R
    =\mathcal{L}_{\mathrm{fuse}}
    +\lambda_u(e)\mathcal{L}_{\mathrm{util}}
    +\lambda_{\mathrm{cons}}\mathcal{L}_{\mathrm{cons}}.
\label{eq:router_objective}
\end{equation}
Both segmentation branches remain frozen during this stage, keeping the predictors fixed while the router learns their relative utility. Fused semantic supervision evaluates the combined prediction, utility learning supplies correction-specific guidance at disagreement pixels, and cross-view consistency encourages stable gate responses under appearance perturbations. Ground-truth category probabilities are used only to construct supervision targets; inference relies entirely on the branch predictions.

\subsection{Optimization and Training Strategy}
\label{subsec:training}

Training proceeds in three sequential stages that separate semantic adaptation, complementary spatial learning, and reliability estimation. First, TCSB undergoes source-domain probing, the rank allocation in Eq.~\eqref{eq:rank_allocation}, and the SVD-based conversion in Eq.~\eqref{eq:svd_factorization}. It is then adapted with the fixed rank map using supervised cross-entropy and the cosine residual-scale schedule.

Next, TCSB is frozen and provides a fixed semantic reference for TSDB training. Only TSDB's adaptation, projection, structure, and decoding modules are updated using Eq.~\eqref{eq:tsdb_loss}. This keeps the basis for complementary supervision fixed while the spatial branch learns. Finally, both segmentation branches are frozen and the router is optimized with Eq.~\eqref{eq:router_objective}, learning to combine the completed predictors without changing their representations.

At inference, the two branches process the unseen image and the router combines their predictions through Eq.~\eqref{eq:bounded_fusion}. Each pixel is assigned the category with the highest fused probability. The complete training procedure uses only source-domain supervision.

\section{Experiments}
\label{sec:experiments}

\subsection{Experimental Setup}
\label{subsec:experimental_setup}

\paragraph{Datasets and transfer settings.}
We evaluate source-only domain generalization on three remote-sensing semantic
segmentation benchmarks. For the ISPRS setting, Potsdam IRRG, Potsdam RGB,
and Vaihingen IRRG are denoted by $P(i)$, $P(r)$, and $V$, respectively. We
consider all six directed transfers among these domains. The resulting
evaluation covers an unseen geographic region, an unseen spectral
configuration, and their compound shift. The six-class label space contains
impervious surface, building, low vegetation, tree, car, and
clutter/background. For LoveDA~\citep{wang2021loveda}, Urban and Rural are
treated as two distinct domains, yielding the $U\to{}R$ and
$R\to{}U$ settings. We additionally evaluate
$U\to\mathrm{OEM}$, in which a model trained on LoveDA Urban is
directly transferred to OpenEarthMap. LoveDA and OpenEarthMap use an aligned
seven-class label space consisting of background, building, road, water,
barren land, forest, and agriculture. All target domains remain unseen during
training.

\begin{table*}[!t]
  \centering
  \caption{Datasets and protocols for the nine source-only DG transfers.}
  \label{tab:dataset_protocols}
  \small
  \setlength{\tabcolsep}{3pt}
  \renewcommand{\arraystretch}{1.12}
  \begin{tabular*}{\textwidth}{@{\extracolsep{\fill}}lllcrrcc@{}}
    \toprule
    Domain shift & Source domain & Target domain & Task
      & Train & Eval. & Image size & Classes \\
    \midrule
    Geographic
      & Potsdam IRRG & Vaihingen IRRG & P(i)2V
      & 3456 & 398 & $512\!\times\!512$ & 6 \\
    Geographic
      & Vaihingen IRRG & Potsdam IRRG & V2P(i)
      & 344 & 2016 & $512\!\times\!512$ & 6 \\
    \addlinespace[2pt]
    Spectral
      & Potsdam RGB & Potsdam IRRG & P(r)2P(i)
      & 3456 & 2016 & $512\!\times\!512$ & 6 \\
    Spectral
      & Potsdam IRRG & Potsdam RGB & P(i)2P(r)
      & 3456 & 2016 & $512\!\times\!512$ & 6 \\
    \addlinespace[2pt]
    Geographic + spectral
      & Potsdam RGB & Vaihingen IRRG & P(r)2V
      & 3456 & 398 & $512\!\times\!512$ & 6 \\
    Geographic + spectral
      & Vaihingen IRRG & Potsdam RGB & V2P(r)
      & 344 & 2016 & $512\!\times\!512$ & 6 \\
    \addlinespace[2pt]
    Scene composition
      & LoveDA Urban & LoveDA Rural & U2R
      & 1156 & 992 & $1024\!\times\!1024$ & 7 \\
    Scene composition
      & LoveDA Rural & LoveDA Urban & R2U
      & 1366 & 677 & $1024\!\times\!1024$ & 7 \\
    Cross-dataset
      & LoveDA Urban & OpenEarthMap & U2OEM
      & 1156 & 384 & $1024\!\times\!1024$ & 7 \\
    \bottomrule
  \end{tabular*}
  \vspace{1mm}
  \parbox{0.99\textwidth}{\footnotesize
    Image size refers to prepared images/crops before resizing to
    $512\!\times\!512$ for network input. Both label protocols include
    background. OpenEarthMap uses the matched 384-image subset.}
\end{table*}

\paragraph{Evaluation protocol and metrics.}
For each directed transfer, we train on labeled source-domain images and
compute target-domain class-wise IoU and mIoU over valid pixels. The SAM3
zero-shot baseline employs the same fixed category-name prompts as TCSB
using the pretrained model parameters.

\paragraph{Compared methods.}
We include conventional segmenters (DeepLabV3+~\citep{chen2018encoder} and SegFormer~\citep{xie2021segformer}), domain-adaptation methods (DAFormer~\citep{hoyer2022daformer} and HRDA~\citep{hoyer2022hrda}), and foundation-model approaches (CrossEarth~\citep{gong2025crossearth} and EarthAdapter~\citep{hu2026earth}). Published comparison results are adopted where available; additional baselines follow their respective evaluation configurations. Performance is summarized using class-wise IoU and mIoU on each source-to-target transfer.

\paragraph{Implementation details.}
All images are resized to $512\!\times\!512$. Both branches are initialized
from SAM3 while the original foundation parameters remain frozen. For TCSB,
LoRA modules are inserted into encoder blocks 3, 7, 11, 15, 19, 23, 27, and 31.
A source-domain probe at rank 128 estimates the contribution of individual
blocks, followed by a one-time allocation of a total rank budget of 512
within $[4,256]$. The converted LoRA factors are optimized using a learning
rate of $5\!\times\!10^{-5}$. TSDB is optimized using AdamW with a learning
rate of $1\!\times\!10^{-4}$ and weight decay of 0.05. After both branches
are frozen, the reliability router is optimized at $2\!\times\!10^{-4}$
with maximum TSDB correction weight $\beta=0.2$. Training and runtime
measurements are conducted on an NVIDIA GeForce RTX 4090 D GPU.

\subsection{Comparison with State-of-the-Art Methods}
\label{subsec:sota_comparison}

We compare ASCoR with representative segmentation methods across nine transfers on the ISPRS, LoveDA, and OpenEarthMap benchmark datasets.
Figure~\ref{fig:overall_radar} summarizes the overall performance,
while Tables~\ref{tab:isprs_cross_domain_full}
and~\ref{tab:loveda_oem_cross_domain_full} report the class-wise results.

\par\medskip\noindent
\begin{minipage}{\columnwidth}
  \centering
  \includegraphics[width=\linewidth]{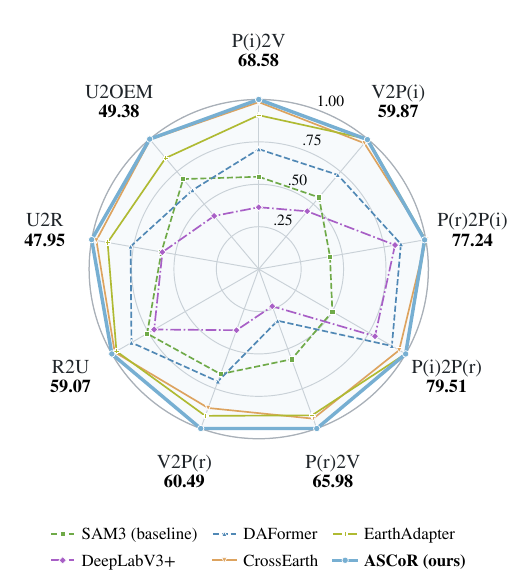}
  \captionof{figure}{Overall performance comparison across nine
  source-only DG transfers on ISPRS, LoveDA, and OpenEarthMap.
  Each axis is normalized by the highest mIoU among the compared
  methods for that task; numerical annotations report the original
  mIoU (\%).}
  \label{fig:overall_radar}
\end{minipage}
\par\medskip



\begin{table*}[!t]
  \centering
  \caption{Per-class IoU and mIoU (\%) on six ISPRS transfers.
  Bold and underlined values indicate the best and second-best results.
  Surf: impervious surface; Bldg: building; Vegt: low vegetation;
  Clut: clutter/background. SAM3 denotes the zero-shot baseline.}
  \label{tab:isprs_cross_domain_full}

  \footnotesize
  \setlength{\tabcolsep}{1pt}
  \renewcommand{\arraystretch}{1.12}

  \resizebox{\textwidth}{!}{%
  \begin{tabular}{@{}
    l@{\hspace{5pt}}c
    @{\hspace{4pt}\vrule width 0.5pt\hspace{4pt}}
    c
    @{\hspace{4pt}\vrule width 0.5pt\hspace{4pt}}
    *{7}{c}
    @{\hspace{4pt}\vrule width 0.5pt\hspace{4pt}}
    c
    @{\hspace{4pt}\vrule width 0.5pt\hspace{4pt}}
    *{7}{c}@{}}

    \toprule
    Method
      & \multicolumn{1}
        {@{}c@{\hspace{4pt}\kern0.5pt\hspace{4pt}}}
        {Backbone}
      & \multicolumn{1}
        {@{}c@{\hspace{4pt}\kern0.5pt\hspace{4pt}}}
        {Task}
      & Surf & Bldg & Vegt & Tree & Car & Clut
      & \multicolumn{1}
        {c@{\hspace{4pt}\kern0.5pt\hspace{4pt}}}
        {mIoU}
      & \multicolumn{1}
        {@{}c@{\hspace{4pt}\kern0.5pt\hspace{4pt}}}
        {Task}
      & Surf & Bldg & Vegt & Tree & Car & Clut & mIoU \\
    \midrule

    SAM3 (baseline) & SAM3
      & \multirow{8}{*}{P(i)2V}
      & 9.47 & 83.46 & 20.32 & 48.36 & \underline{60.25} & 2.12 & 37.33
      & \multirow{8}{*}{V2P(i)}
      & 17.17 & 70.15 & 9.37 & 35.27 & 62.51 & 4.51 & 33.17 \\

    DeepLabV3+ & R101
      & & 9.15 & 60.94 & 17.60 & 52.28 & 7.61 & 2.54 & 25.02
      & & 29.32 & 39.45 & 29.17 & 24.92 & 37.42 & 0.00 & 26.72 \\

    SegFormer & MiT-B5
      & & 60.58 & 74.72 & 27.68 & 63.69 & 33.38 & 7.56 & 44.60
      & & 66.01 & 70.94 & 48.53 & 13.19 & 65.98 & 3.00 & 44.61 \\

    DAFormer & MiT-B5
      & & 61.48 & 75.53 & 42.87 & 66.57 & 34.52 & 9.99 & 48.49
      & & 62.22 & 70.86 & 43.09 & 13.86 & 64.31 & \underline{6.95} & 43.55 \\

    HRDA & MiT-B5
      & & 64.14 & 76.65 & 42.06 & 65.11 & 28.72 & 13.52 & 48.37
      & & 68.64 & 77.24 & 46.33 & 11.05 & 64.12 & 6.48 & 45.64 \\

    CrossEarth & ViT-L
      & & \textbf{83.60} & \textbf{91.70} & \underline{63.80}
      & 70.20 & 59.50 & \textbf{36.30} & \underline{67.50}
      & & \textbf{77.10} & 80.80 & \underline{61.30}
      & 38.90 & \underline{82.90} & \textbf{8.40} & 58.20 \\

    EarthAdapter & DINOv2-L
      & & 77.28 & 88.16 & 59.37 & \underline{71.71}
      & 58.01 & 18.64 & 62.20
      & & \underline{76.21} & \textbf{86.98} & \textbf{63.56}
      & \underline{45.27} & \textbf{86.33} & 2.16 & \textbf{60.08} \\

    \textbf{ASCoR (ours)} & SAM3
      & & \underline{80.87} & \underline{88.57} & \textbf{65.15}
      & \textbf{74.66} & \textbf{69.13} & \underline{33.10} & \textbf{68.58}
      & & 72.81 & \underline{85.04} & 54.81
      & \textbf{60.29} & 81.74 & 4.54 & \underline{59.87} \\

    \midrule

    SAM3 (baseline) & SAM3
      & \multirow{8}{*}{P(r)2P(i)}
      & 17.17 & 70.15 & 9.37 & 35.27 & 62.51 & 4.51 & 33.17
      & \multirow{8}{*}{P(i)2P(r)}
      & 27.22 & 75.39 & 31.86 & 33.39 & 66.51 & 4.58 & 39.83 \\

    DeepLabV3+ & R101
      & & 73.68 & 76.42 & 46.36 & 71.00 & 87.87 & 25.44 & 63.46
      & & 66.95 & 81.76 & 52.62 & 69.82 & 88.01 & 18.29 & 62.91 \\

    SegFormer & MiT-B5
      & & 82.12 & 89.98 & 51.51 & 72.27 & 89.69 & 27.35 & 68.82
      & & 80.22 & 92.38 & 56.55 & 56.28 & 89.56 & 19.50 & 65.75 \\

    DAFormer & MiT-B5
      & & 82.06 & 89.03 & 44.03 & 71.77 & 88.05 & 22.21 & 66.19
      & & 81.95 & 92.06 & 67.02 & 68.85 & 89.84 & 32.54 & 72.04 \\

    HRDA & MiT-B5
      & & 81.94 & 84.24 & 36.89 & 70.07 & 89.69 & 18.97 & 63.63
      & & 80.59 & 90.97 & 66.14 & 70.85 & 88.58 & 31.63 & 71.46 \\

    CrossEarth & ViT-L
      & & 86.10 & \underline{91.50} & \textbf{73.80}
      & \underline{79.70} & 91.90 & \textbf{43.80} & \textbf{77.80}
      & & 86.30 & 93.50 & 73.80 & 74.00 & 91.20 & 37.20 & 76.00 \\

    EarthAdapter & DINOv2-L
      & & \underline{86.40} & 90.58 & \underline{73.01}
      & \textbf{80.30} & \underline{93.13} & \underline{42.06} & \underline{77.58}
      & & \textbf{87.80} & \textbf{94.32} & \textbf{78.29}
      & \underline{78.26} & \underline{93.27} & \textbf{44.85} & \underline{79.46} \\

    \textbf{ASCoR (ours)} & SAM3
      & & \textbf{86.47} & \textbf{92.36} & 72.16
      & 78.98 & \textbf{94.49} & 38.95 & 77.24
      & & \underline{87.71} & \underline{94.29} & \underline{78.24}
      & \textbf{78.71} & \textbf{93.75} & \underline{44.38} & \textbf{79.51} \\

    \midrule

    SAM3 (baseline) & SAM3
      & \multirow{8}{*}{P(r)2V}
      & 9.47 & 83.46 & 20.32 & 48.36 & 60.25 & 2.12 & 37.33
      & \multirow{8}{*}{V2P(r)}
      & 27.22 & 75.39 & 31.86 & \underline{33.39} & 66.51 & 4.58 & 39.83 \\

    DeepLabV3+ & R101
      & & 5.11 & 52.13 & 7.36 & 25.28 & 0.79 & 1.60 & 15.38
      & & 23.33 & 34.93 & 25.18 & 25.15 & 30.52 & 0.00 & 23.18 \\

    SegFormer & MiT-B5
      & & 48.83 & 67.72 & 1.84 & 23.62 & 28.82 & 1.58 & 28.73
      & & 54.44 & 58.86 & 35.00 & 2.05 & 67.31 & 0.98 & 36.44 \\

    DAFormer & MiT-B5
      & & 40.59 & 57.49 & 0.29 & 4.59 & 23.97 & 1.30 & 21.37
      & & 59.25 & 73.10 & 43.86 & 6.90 & 66.00 & \underline{6.64} & 42.63 \\

    HRDA & MiT-B5
      & & 44.41 & 61.92 & 3.36 & 3.87 & 24.67 & 1.46 & 23.28
      & & 55.26 & 70.42 & 36.34 & 3.55 & 65.78 & 3.79 & 39.19 \\

    CrossEarth & ViT-L
      & & \underline{78.10} & \textbf{89.30} & \underline{55.20}
      & \underline{72.50} & \underline{61.90} & 14.50 & \underline{61.90}
      & & \underline{72.50} & 73.60 & 58.70
      & 22.40 & \underline{81.10} & \textbf{7.60} & 52.70 \\

    EarthAdapter & DINOv2-L
      & & 76.25 & 86.49 & 51.15 & \textbf{73.26}
      & 58.84 & \underline{17.13} & 60.52
      & & 71.05 & \underline{82.73} & \underline{61.18}
      & 32.76 & \textbf{84.57} & 1.83 & \underline{55.69} \\

    \textbf{ASCoR (ours)} & SAM3
      & & \textbf{79.74} & \underline{88.86} & \textbf{62.43}
      & 69.61 & \textbf{68.36} & \textbf{26.88} & \textbf{65.98}
      & & \textbf{74.71} & \textbf{85.93} & \textbf{64.21}
      & \textbf{54.83} & 79.72 & 3.56 & \textbf{60.49} \\

    \bottomrule
  \end{tabular}%
  }
\end{table*}


\begin{table}[!tbp]
  \centering
  \caption{Per-class IoU and mIoU (\%) on LoveDA and OpenEarthMap.
  Bold and underlined values rank first and second within each transfer.
  Bkgd: background; Bldg: building; Rd: road; Wtr: water;
  Barr: barren; Frst: forest; Agri: agriculture.}
  \label{tab:loveda_oem_cross_domain_full}

  \setlength{\tabcolsep}{2.6pt}
  \renewcommand{\arraystretch}{1.14}

  \resizebox{\columnwidth}{!}{%
  \begin{tabular}{l@{\hspace{5pt}\vrule width .35pt\hspace{5pt}}*{7}{c}@{\hspace{6pt}\vrule width .35pt\hspace{6pt}}c}
    \toprule
    Method & Bkgd & Bldg & Rd & Wtr & Barr & Frst & Agri & mIoU (\%) \\
    \addlinespace[5pt]
    \midrule

    \multicolumn{9}{c}{Performance comparison: LoveDA Rural $\rightarrow$ LoveDA Urban} \\
    \midrule

    SAM3 (baseline)
      & 17.87 & 53.01 & 55.71 & 58.74 & 34.85 & 50.69 & 43.31 & 44.88 \\
    DeepLabV3+ R101
      & 34.56 & 48.35 & 38.70 & 56.61 & 34.02 & 36.27 & 45.92 & 42.06 \\
    SegFormer-B5
      & 34.06 & 52.63 & 54.30 & 71.79 & 43.53 & 38.81 & 52.77 & 49.70 \\
    DAFormer
      & 35.49 & 52.74 & 51.51 & 73.10 & 45.03 & 47.13 & 53.20 & 51.17 \\
    HRDA
      & 38.80 & 51.19 & 48.06 & 72.47 & 46.01 & 34.60 & 56.65 & 49.68 \\
    CrossEarth
      & 39.80 & \underline{63.30} & 57.30 & \textbf{75.90}
      & \underline{51.80} & \underline{55.00} & \textbf{62.50} & \underline{57.90} \\
    EarthAdapter
      & \underline{42.16} & 60.19 & \underline{58.12} & \underline{75.21}
      & \textbf{53.08} & 50.09 & 61.64 & 57.21 \\
    \textbf{ASCoR (ours)}
      & \textbf{42.63} & \textbf{65.87} & \textbf{58.30} & 74.70
      & 51.62 & \textbf{58.34} & \underline{62.04} & \textbf{59.07} \\

    \addlinespace[5pt]
    \midrule
    \multicolumn{9}{c}{Performance comparison: LoveDA Urban $\rightarrow$ LoveDA Rural} \\
    \midrule

    SAM3 (baseline)
      & 2.27 & 36.29 & 43.41 & 53.12 & 10.68 & 13.68 & 36.10 & 27.94 \\
    DeepLabV3+ R101
      & 46.92 & 27.89 & 29.93 & 30.14 & 9.45 & 19.43 & 29.69 & 27.64 \\
    SegFormer-B5
      & \underline{57.02} & 47.55 & 37.13 & 64.56 & 8.86 & 3.34 & 47.78 & 38.03 \\
    DAFormer
      & 49.29 & 43.45 & 33.17 & 64.52 & 4.38 & 14.48 & 48.28 & 36.79 \\
    HRDA
      & 55.27 & 32.09 & 36.48 & 59.86 & 6.67 & 18.26 & 52.07 & 37.24 \\
    CrossEarth
      & \textbf{57.50} & \textbf{61.50} & \underline{48.50} & 65.60
      & 9.60 & 25.40 & \textbf{58.20} & \underline{46.60} \\
    EarthAdapter
      & 33.61 & 53.00 & 44.51 & \textbf{68.44}
      & \underline{16.44} & \underline{35.01} & 52.67 & 43.38 \\
    \textbf{ASCoR (ours)}
      & 53.40 & \underline{57.76} & \textbf{49.62} & \underline{66.85}
      & \textbf{17.38} & \textbf{35.82} & \underline{54.79} & \textbf{47.95} \\

    \addlinespace[5pt]
    \midrule
    \multicolumn{9}{c}{Performance comparison: LoveDA Urban $\rightarrow$ OpenEarthMap} \\
    \midrule

    SAM3 (baseline)
      & 7.98 & 48.38 & \textbf{46.96} & 66.59
      & \underline{9.53} & 42.00 & 18.08 & 34.22 \\
    DeepLabV3+ R101
      & 41.00 & 26.71 & 22.73 & 18.14 & 6.78 & 13.39 & 12.59 & 20.19 \\
    SegFormer-B5
      & 35.81 & 45.16 & 26.42 & 18.22 & 2.58 & 30.97 & 22.16 & 25.90 \\
    DAFormer
      & 35.06 & 45.18 & 25.26 & 33.45 & 4.91 & 36.90 & 29.26 & 30.00 \\
    HRDA
      & 41.99 & 44.03 & 20.57 & 33.72 & 5.62 & 31.93 & 32.09 & 29.99 \\
    CrossEarth
      & \underline{47.60} & \underline{65.10} & 39.90 & \underline{69.90}
      & \textbf{13.20} & 42.40 & \textbf{66.30} & \underline{49.20} \\
    EarthAdapter
      & 41.71 & 51.32 & 25.25 & 63.04
      & 6.63 & \underline{45.89} & \underline{61.51} & 42.19 \\
    \textbf{ASCoR (ours)}
      & \textbf{49.51} & \textbf{70.02} & \underline{45.44} & \textbf{70.44}
      & 4.98 & \textbf{51.81} & 53.46 & \textbf{49.38} \\

    \bottomrule
  \end{tabular}%
  }
\end{table}

\paragraph{Results on ISPRS.}
As reported in Table~\ref{tab:isprs_cross_domain_full}, ASCoR achieves
the highest mIoU in four of the six directed ISPRS transfers, with an
average of 68.61\%. This exceeds the 65.92\% mean of EarthAdapter and
the 65.68\% mean of CrossEarth. The strongest gains emerge under
combined geographic and spectral shifts: ASCoR obtains 65.98\% on
$P(r)\to{}V$ and 60.49\% on $V\to{}P(r)$, improving on EarthAdapter
by 5.46 and 4.80 points, respectively. For $P(r)\to{}V$, low vegetation,
car, and clutter reach 62.43\%, 68.36\%, and 26.88\% IoU. On the reverse
$V\to{}P(r)$ transfer, tree IoU rises to 54.83\%, compared with 32.76\%
for EarthAdapter. ASCoR also leads $P(i)\to{}V$ and $P(i)\to{}P(r)$.
In the remaining two directions, the gaps to the best reported mIoU
are just 0.21 points on $V\to{}P(i)$ and 0.56 points on
$P(r)\to{}P(i)$. These results show the strength of the learned
representations across geographical, spectral, and compound shifts.

\paragraph{Results on LoveDA and OpenEarthMap.}
Table~\ref{tab:loveda_oem_cross_domain_full} shows the best overall
mIoU in all three transfers: 59.07\% for $R\to{}U$, 47.95\% for
$U\to{}R$, and 49.38\% for $U\to\mathrm{OEM}$. On $R\to{}U$,
ASCoR leads in building (65.87\%), forest (58.34\%), and road (58.30\%)
segmentation. On $U\to{}R$, the improvements are especially clear for
barren land (17.38\%) and forest (35.82\%). The cross-dataset
$U\to\mathrm{OEM}$ transfer further achieves 70.02\% building IoU,
70.44\% water IoU, and 45.44\% road IoU, demonstrating strong semantic
and structural transfer to imagery collected under a different dataset
protocol. The results cover both urban--rural scene changes and the more
challenging cross-dataset setting.

\subsection{Component Generalization across Domain Shifts}
\label{subsec:component_generalization}

We examine the contributions of the three components under representative scene-composition (\(U\to{}R\)), spectral (\(P(r)\to{}P(i)\)), and joint geographic--spectral (\(P(r)\to{}V\)) changes. Uniform LoRA employs the same eight adapted SAM3 blocks and the same rank budget of 512 as TCSB, with rank 64 in each block. The uniform- and adaptive-rank variants therefore evaluate the allocation strategy under a matched backbone and adaptation budget. Likewise, the fixed- and learned-fusion variants use identical trained branches, isolating the contribution of reliability-guided correction from changes in backbone capacity.

\begin{table}[!t]
  \centering
  \caption{Component ablation across three representative transfers.
  Values are mIoU (\%).}
  \label{tab:cross_setting_component_ablation}

  \fontsize{8.5}{10}\selectfont
  \setlength{\tabcolsep}{1.5pt}
  \renewcommand{\arraystretch}{1.25}

  \setlength{\aboverulesep}{0pt}
  \setlength{\belowrulesep}{0pt}

  \begin{tabular*}{\columnwidth}{@{\extracolsep{\fill}}
    cccc
    @{\hspace{3pt}\vrule width 0.5pt\hspace{3pt}}
    cccc@{}}

    \toprule
    TCSB & AR & TSDB & Router
      & U2R & P(r)2P(i) & P(r)2V & Avg. \\
    \midrule

    -- & -- & -- & --
      & 27.94 & 33.17 & 37.33 & 32.81 \\

    $\checkmark$ & -- & -- & --
      & 45.65 & 76.78 & 65.10 & 62.51 \\

    $\checkmark$ & $\checkmark$ & -- & --
      & 46.88 & 76.88 & 64.69 & 62.82 \\

    -- & -- & $\checkmark$ & --
      & 35.01 & 69.98 & 51.74 & 52.24 \\

    $\checkmark$ & $\checkmark$ & $\checkmark$ & --
      & \underline{47.45} & \underline{77.07}
      & \underline{65.37} & \underline{63.30} \\

    $\checkmark$ & $\checkmark$ & $\checkmark$ & $\checkmark$
      & \textbf{47.95} & \textbf{77.24}
      & \textbf{65.98} & \textbf{63.72} \\

    \bottomrule
  \end{tabular*}

  \vspace{2pt}
  \parbox{\columnwidth}{\footnotesize
    AR: probe-once adaptive rank allocation in TCSB. Without AR,
    TCSB uses uniform rank 64. The all-dash row denotes zero-shot
    SAM3. Dual-branch variants without the router use a fixed
    TSDB weight of 0.1.}
\end{table}

Table~\ref{tab:cross_setting_component_ablation} shows how the
components contribute across different transfers. Adaptive rank allocation
raises the three-task mean from 62.51\% with uniform LoRA to 62.82\%
with TCSB. Combining the dense predictions of TSDB with TCSB increases accuracy
in all three settings, highlighting complementary information across the branches. Adding
the reliability router yields further improvements of 0.50, 0.17, and
0.61 points over fixed fusion, giving the complete framework the highest
mean of 63.72\%. The results demonstrate that source-guided dense
features and spatial reliability cues provide complementary benefits to
the category-conditioned semantic predictions.
\begin{table*}[!t]
  \centering
  \caption{Class-wise component ablation on LoveDA Urban
  $\rightarrow$ Rural. Values are IoU and mIoU (\%).}
  \label{tab:u2r_classwise_ablation}

  \small
  \setlength{\tabcolsep}{3pt}
  \renewcommand{\arraystretch}{1.25}
  \setlength{\aboverulesep}{0pt}
  \setlength{\belowrulesep}{0pt}

  \begin{tabular*}{\textwidth}{@{\extracolsep{\fill}}
    cccc
    @{\hspace{4pt}\vrule width 0.5pt\hspace{4pt}}
    *{8}{c}@{}}

    \toprule
    TCSB & AR & TSDB & Router
      & Bkgd & Bldg & Road & Water
      & Barren & Forest & Agri & mIoU \\
    \midrule

    -- & -- & -- & --
      & 2.27 & 36.29 & 43.41 & 53.12
      & 10.68 & 13.68 & 36.10 & 27.94 \\

    $\checkmark$ & -- & -- & --
      & 54.04 & 60.53 & 49.46 & 63.86
      & 11.05 & 24.43 & 56.16 & 45.65 \\

    $\checkmark$ & $\checkmark$ & -- & --
      & 49.53 & 55.23 & 48.57 & 66.02
      & 17.71 & 35.76 & 55.35 & 46.88 \\

    -- & -- & $\checkmark$ & --
      & 52.06 & 46.33 & 36.76 & 61.61
      & 12.91 & 7.38 & 28.02 & 35.01 \\

    $\checkmark$ & $\checkmark$ & $\checkmark$ & --
      & 51.50 & 56.46 & 49.07 & 66.41
      & 17.53 & 35.91 & 55.28 & \underline{47.45} \\

    $\checkmark$ & $\checkmark$ & $\checkmark$ & $\checkmark$
      & 53.41 & 57.75 & 49.62 & 66.85
      & 17.38 & 35.82 & 54.79 & \textbf{47.95} \\

    \bottomrule
  \end{tabular*}

  \vspace{2pt}
  \parbox{\textwidth}{\footnotesize
    AR: probe-once adaptive rank allocation in TCSB. Without AR,
    TCSB uses uniform rank 64. The all-dash row denotes zero-shot
    SAM3. Dual-branch variants without the router use a fixed
    TSDB weight of 0.1.}
\end{table*}
\subsection{Class-wise Complementarity Analysis}
\label{subsec:classwise_complementarity}

To understand how the weaker auxiliary TSDB can still improve the primary
branch, we report the complete class-wise decomposition on
$U\to{}R$ in Table~\ref{tab:u2r_classwise_ablation}. This transfer
contains substantial changes in scene composition and object appearance and
therefore provides a suitable diagnostic setting.

Uniform fixed-rank adaptation improves the zero-shot result by 17.71 mIoU
points. Dynamic allocation further increases the TCSB score by 1.23 points,
with marked improvements for water, barren land, and forest. TSDB provides
additional evidence for background, buildings, and water. The fixed
combination reaches 47.45\%, and reliability-guided fusion increases mIoU
to 47.95\%. This progression illustrates how the two branches jointly
recover difficult categories in the $U\to{}R$ transfer.

\subsection{Analysis of the Reliability-Router Objective}
\label{subsec:router_objective_ablation}

We next isolate the optimization terms used to learn the reliability router.
Both branches remain frozen in this experiment, isolating the contribution
of the router objective.

\begin{table}[!tbp]
  \centering
  \caption{Ablation of the reliability-router objective on LoveDA
  Urban $\rightarrow$ Rural. Both branches and all inference settings
  remain fixed.}
  \label{tab:router_loss_ablation}

  \small
  \setlength{\tabcolsep}{3pt}
  \renewcommand{\arraystretch}{1.25}
  \setlength{\aboverulesep}{0pt}
  \setlength{\belowrulesep}{0pt}

  \begin{tabular*}{\columnwidth}{@{\extracolsep{\fill}}
    ccc
    @{\hspace{4pt}\vrule width 0.5pt\hspace{4pt}}
    cc@{}}

    \toprule
    $\mathcal{L}_{\mathrm{fuse}}$
      & $\mathcal{L}_{\mathrm{util}}$
      & $\mathcal{L}_{\mathrm{cons}}$
      & mIoU (\%) & Gain (pp) \\
    \midrule

    $\checkmark$ & -- & --
      & 47.93 & -- \\

    $\checkmark$ & $\checkmark$ & --
      & \underline{47.94} & +0.01 \\

    $\checkmark$ & $\checkmark$ & $\checkmark$
      & \textbf{47.95} & +0.02 \\

    \bottomrule
  \end{tabular*}
\end{table}

Table~\ref{tab:router_loss_ablation} isolates the effect of the
router objective with both branches fixed. Fused semantic supervision
achieves 47.93\% mIoU. Incorporating relative utility increases the
score to 47.94\%, and cross-view consistency further raises it to
47.95\%. These incremental gains complement the larger improvements
from adding TSDB and learning spatially adaptive fusion, as shown in
Table~\ref{tab:cross_setting_component_ablation}.


\par\medskip\noindent\begin{minipage}{\columnwidth}
\centering
\includegraphics[width=0.98\columnwidth]{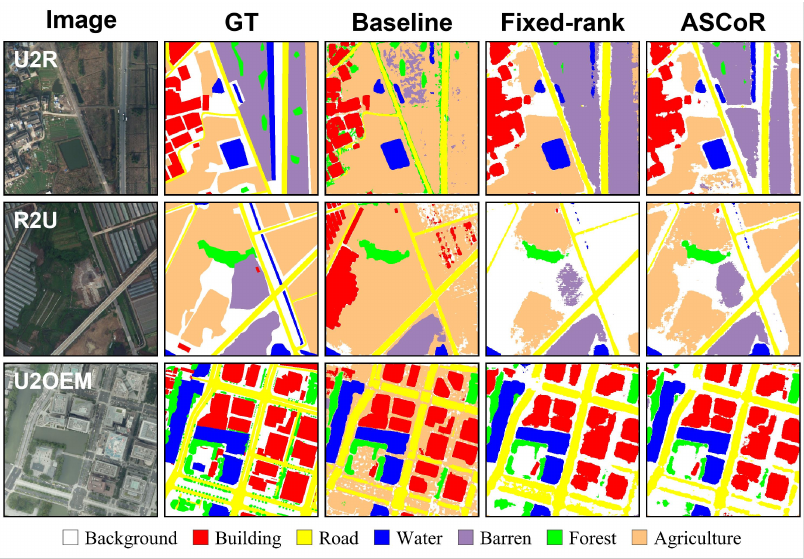}
\captionof{figure}{Qualitative comparison on the R2U, U2R, and U2OEM
transfers. Columns show the input image, ground truth (GT),
zero-shot SAM3, uniform fixed-rank LoRA, and ASCoR, illustrating
segmentation differences under urban--rural and cross-dataset shifts.}
\label{fig:loveda_qualitative}
\end{minipage}\par\medskip

\subsection{Parameter Efficiency}
\label{subsec:parameter_efficiency}

Table~\ref{tab:parameter_efficiency} distinguishes physically allocated
task-specific parameters from the subset that is active under the final rank
configuration. This distinction is necessary because Dynamic Adaptive LoRA
retains maximum-rank tensors for flexible allocation but activates only the
slices selected by the learned rank map.

\begin{table}[!tbp]
  \centering
  \caption{Parameter analysis. ``Trainable'' counts physically allocated
  task-specific parameters, following the conventional PEFT reporting
  protocol. ``Active'' counts parameters participating in the final forward
  computation.}
  \label{tab:parameter_efficiency}
  \setlength{\tabcolsep}{3.4pt}
  \renewcommand{\arraystretch}{1.04}
  \resizebox{\columnwidth}{!}{%
  \begin{tabular}{lrrrr}
    \toprule
    Model/component & Total & Trainable & Ratio & Active \\
                    & (M)   & (M)       & (\%)  & (M) \\
    \midrule
    TCSB               & 853.09  & 12.583 & 1.475 &  3.146 \\
    TSDB               & 868.22  & 27.715 & 3.192 & 27.715 \\
    \textbf{Complete framework}
                       & 1721.32 & 40.302 & 2.341 & 30.865 \\
    \bottomrule
  \end{tabular}%
  }
\end{table}

The complete framework contains 1.721B resident parameters, of which 40.302M (2.341\%) are task-specific. Stage-wise optimization keeps the largest number of simultaneously trainable parameters to 27.715M, equivalent to 1.61\% of the complete model. Router learning updates only 4,033 parameters, and the final forward computation activates 30.865M task-specific parameters (1.793\%). This design concentrates task learning in a small set of parameters while retaining the pretrained representations of both branches. The two SAM3 encoders account for most of the total model size and inference cost.

\subsection{Runtime Overhead of Reliability-Guided Fusion}
\label{subsec:runtime_efficiency}

To isolate the cost of the learned fusion mechanism, we compare it with fixed
dual-branch fusion while keeping both branch forward passes identical. Runtime
is measured at $512\!\times\!512$ resolution with batch size one on an RTX
4090 D. We exclude image-decoding time, perform three warm-up iterations, and
report the median of 20 CUDA-synchronized measurements.

\begin{table}[!tbp]
  \centering
  \caption{Inference latency of fixed and learned dual-branch fusion. The
  reported time includes both branches and the corresponding fusion operation.}
  \label{tab:router_latency}
  \setlength{\tabcolsep}{4.0pt}
  \renewcommand{\arraystretch}{1.04}
  \resizebox{\columnwidth}{!}{%
  \begin{tabular}{lcccc}
    \toprule
    Fusion strategy
      & $P(r)\to{}P(i)$
      & $P(r)\to{}V$
      & Average
      & Overhead \\
    \midrule
    Fixed $w_B=0.1$       & 525.51 & 522.33 & 523.92 & -- \\
    Reliability-guided    & 527.02 & 523.83 & 525.42 & +0.29\% \\
    \bottomrule
  \end{tabular}%
  }
\end{table}

The learned router adds approximately 1.50 ms per image to the complete
dual-branch pipeline, corresponding to 0.29\% overhead relative to fixed
fusion. Most inference time is spent computing the two SAM3 branch
predictions, while reliability-guided correction requires little additional
computation.

\subsection{Qualitative and Diagnostic Analysis}
\label{subsec:qualitative}

\paragraph{Qualitative segmentation results.}
Figure~\ref{fig:loveda_qualitative} compares the unadapted SAM3 baseline,
fixed-rank adaptation, and ASCoR on the three LoveDA/OpenEarthMap transfers.
The baseline shows fragmented land-cover regions and confusion among
visually similar classes. Source-domain adaptation restores more coherent
regions; ASCoR further improves object boundaries and the continuity of
roads, farmland, and built-up structures in these examples.
\begin{figure*}[!t]
  \centering
  \includegraphics[width=\textwidth]{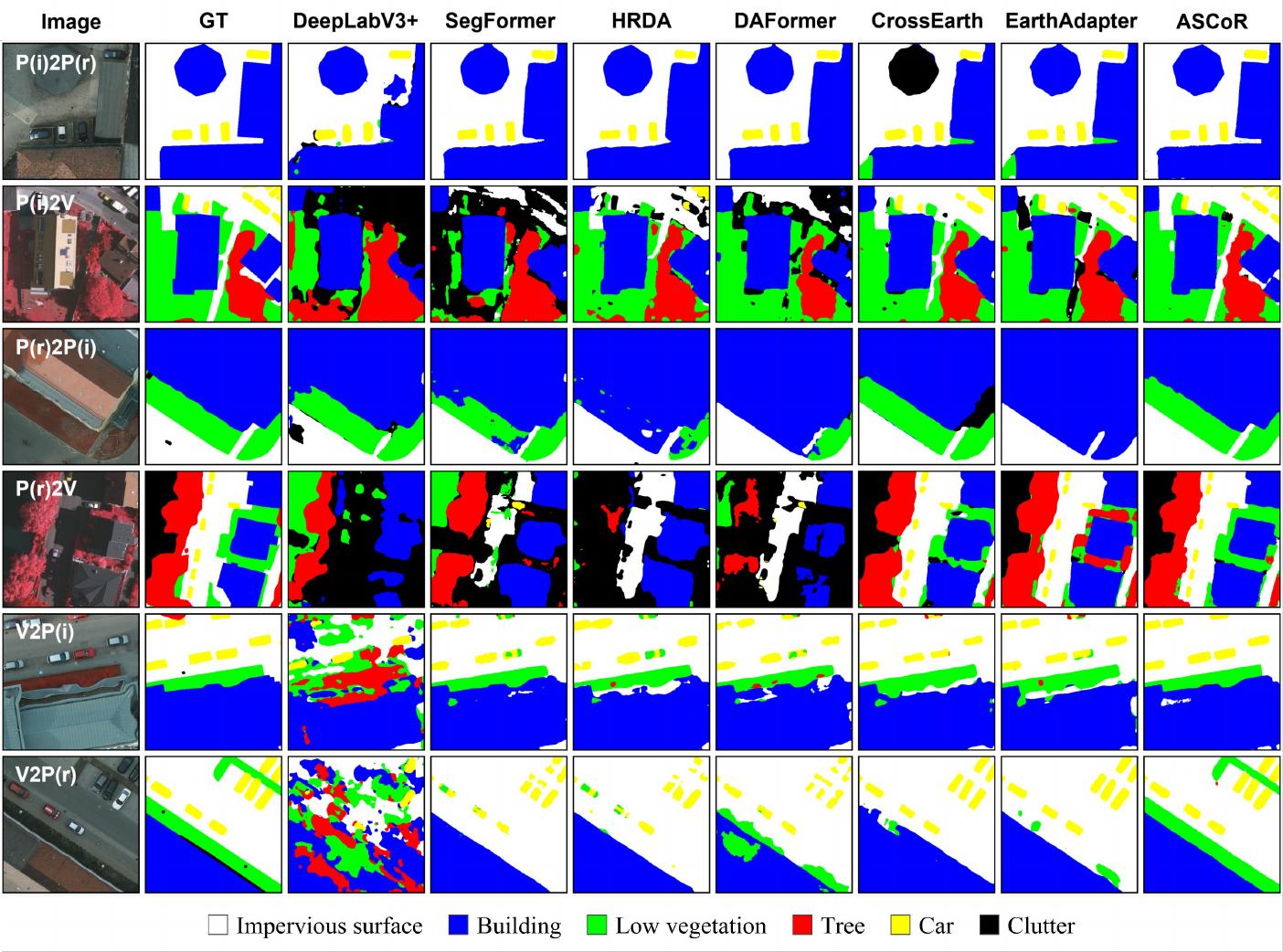}
  \caption{Qualitative comparison across the six directed ISPRS
transfers among Potsdam IRRG, Potsdam RGB, and Vaihingen IRRG.
Columns show the input image, ground truth (GT), DeepLabV3+,
SegFormer, HRDA, DAFormer, CrossEarth, EarthAdapter, and ASCoR,
covering geographic, spectral, and combined domain shifts.}
  \label{fig:isprs_qualitative}
\end{figure*}

Figure~\ref{fig:isprs_qualitative} presents examples from all six ISPRS
transfers. Under combined geographic and spectral changes, conventional
segmentation methods show missing structures and fragmented predictions.
ASCoR maintains more consistent building footprints, surface regions, and
linear boundaries, agreeing with its strong performance on the challenging
$P(r)\to{}V$ and $V\to{}P(r)$ tasks.

\paragraph{Adaptive rank and representation evolution.}
Probe-once allocation assigns more low-rank capacity to later SAM3 encoder blocks, concentrating adaptation where task-specific semantic features emerge. Figure~\ref{fig:r2u_layers} compares the pretrained SAM3 baseline and ASCoR at blocks 3, 15, 23, and 31. Differences are relatively small in shallow responses and become more prominent in deeper blocks, where the adapted representation emphasizes land-cover structures and their boundaries. The deeper difference maps also highlight object interiors, suggesting changes in regional responses beyond boundaries.

\par\medskip\noindent\begin{minipage}{\columnwidth}
\centering
\includegraphics[width=0.98\columnwidth]{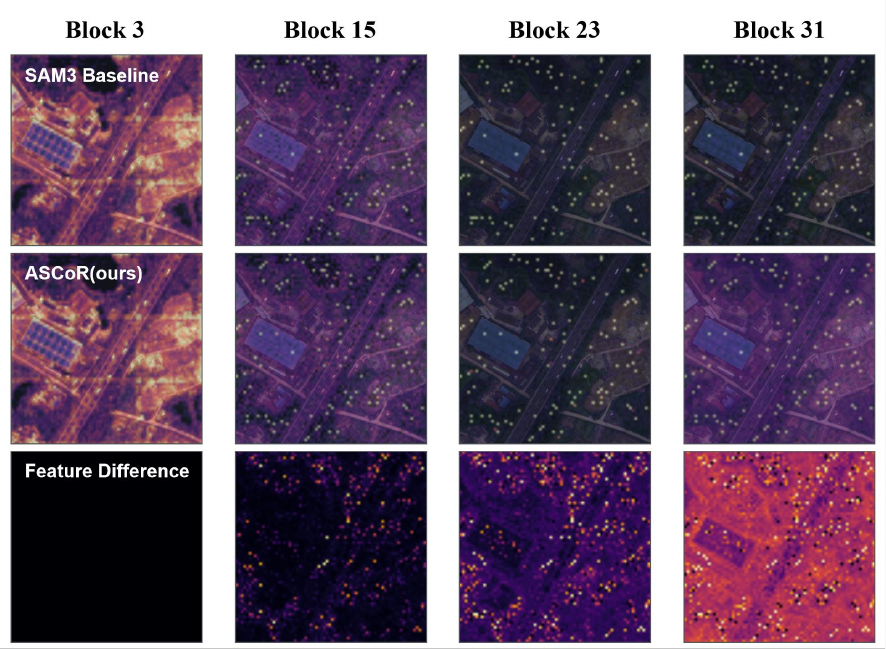}
\captionof{figure}{Feature responses of pretrained SAM3 and ASCoR on
$R\to{}U$ at blocks 3, 15, 23, and 31. The bottom row shows their
differences; warmer colors indicate stronger responses.}
\label{fig:r2u_layers}
\end{minipage}\par\medskip

\paragraph{Complementary correction behavior.}
Figure~\ref{fig:r2u_reliability} illustrates the spatially varying
reliability of the two branches. Their confidence and entropy maps highlight
different regions of uncertainty, while the disagreement map identifies
locations where dense predictions can rescue errors from TCSB. 

\par\medskip\noindent
\begin{minipage}{\columnwidth}
{\centering
\includegraphics[width=0.98\linewidth]{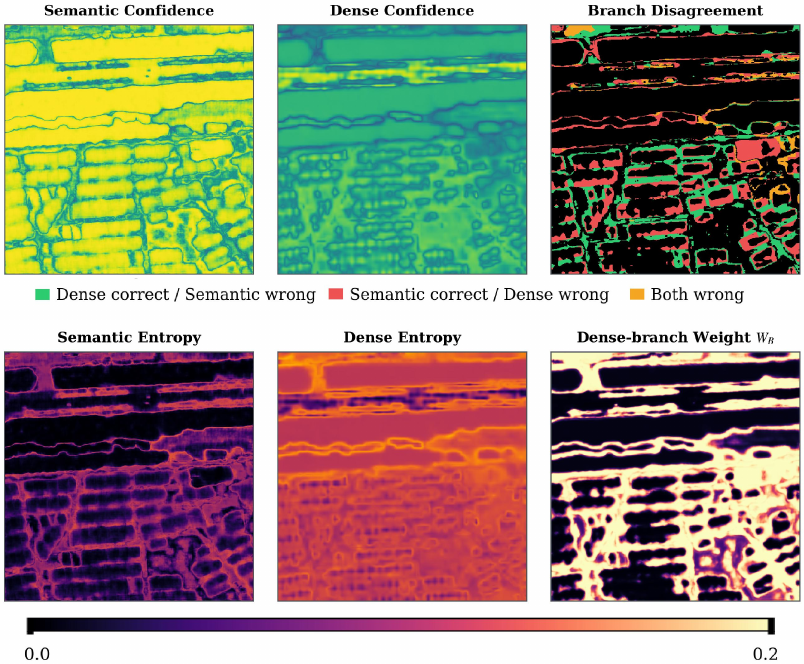}
\captionof{figure}{Spatial reliability analysis on the R2U transfer, showing branch-wise confidence and entropy, inter-branch disagreement, and learned TSDB correction weights.}
\label{fig:r2u_reliability}
\par}
\smallskip
\noindent The router assigns higher TSDB weights around selected boundaries and complex structures. This pattern reflects spatially varying contributions from the dense branch.
\par
\end{minipage}
\par\medskip

\par\medskip\noindent\begin{minipage}{\columnwidth}
  \centering
  \includegraphics[width=\columnwidth]{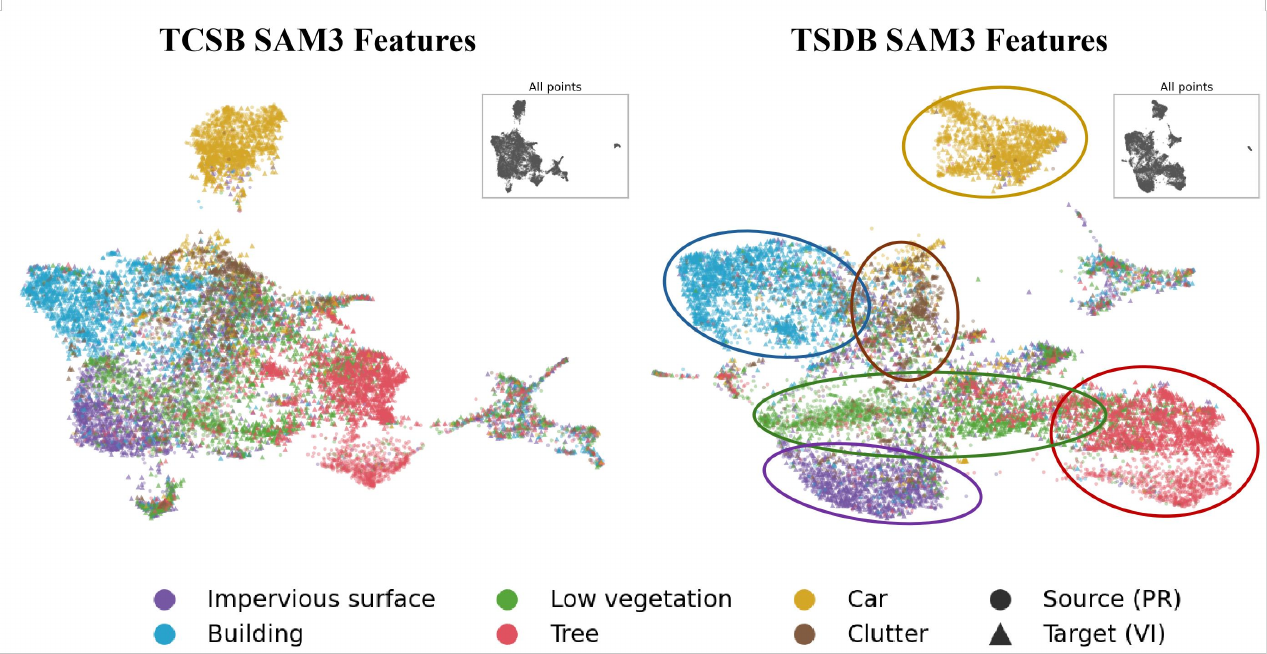}
  \captionof{figure}{UMAP of TCSB (left) and TSDB (right) on
  $P(r)\to{}V$. Colors indicate semantic classes; circles and triangles
  denote source and target samples, respectively.}
  \label{fig:umap_features}
\end{minipage}\par
\makeatletter
\@startsection{paragraph}{4}{\parindent}
  {2pt}{-6pt}
  {\ssssparaindent\ssssectionfont\itshape\raggedright}
  {Feature-space visualization.}
\makeatother
Figure~\ref{fig:umap_features} presents UMAP projections of TCSB and TSDB
features for the $P(r)\to{}V$ transfer. Both branches preserve recognizable
semantic groupings across geographic and spectral changes: building features
occupy a broad region, trees extend toward a separate region, and cars form
an especially compact cluster. Source and target samples overlap within
several category groups, reflecting transferable category organization.
The differences between the two feature arrangements further reveal how
category-aware semantics and dense spatial modeling provide distinct yet
complementary representations, consistent with the fusion improvements in
Table~\ref{tab:cross_setting_component_ablation}.


\section{Conclusion}
\label{sec:conclusion}

We introduced ASCoR, an asymmetric dual-branch SAM3 framework for source-only
domain generalization in remote-sensing semantic segmentation. TCSB uses
category-conditioned SAM3 predictions and Probe-Once Dynamic Adaptive LoRA
to preserve transferable semantics under a constrained adaptation budget.
TSDB learns dense multi-scale and structural features from a separate frozen
foundation encoder. A lightweight reliability router combines their
complementary predictions through spatially adaptive correction.

Across six ISPRS transfers, ASCoR achieves 68.61\% average mIoU and ranks
first on four directions, including the challenging joint geographic--spectral
transfers. The two remaining directions are within 0.56 percentage points
of the best results. ASCoR also attains the highest mIoU on all three
LoveDA/OpenEarthMap transfers, reaching 59.07\%, 47.95\%, and 49.38\%.
Component studies highlight the gains from adaptive rank allocation,
complementary dense representations, and reliability-guided fusion.
The framework updates at most 1.61\% of its resident parameters during
any training stage, while router inference adds only 0.29\% latency.
These findings establish asymmetric foundation-model adaptation as an
effective strategy for transferable dense land-cover understanding.

\printcredits
\section*{Acknowledgments}

This work was supported by the Natural Science Foundation of Sichuan Province (Grant No. 2026NSFSC0223).
\bibliographystyle{cas-model2-names}
\setlength{\bibsep}{0pt plus 1pt}
\bibliography{refs}

\end{document}